\documentclass{article} 
\usepackage{nips13submit_e,times}
\usepackage{hyperref}
\usepackage{graphicx}
\usepackage{url}

\title{Deep Learning for Music}

\author{
Allen Huang \\
Department of Management Science and Engineering\\
Stanford University\\
\texttt{allenh@cs.stanford.edu} \\
\And
Raymond Wu \\
Department of Computer Science \\
Stanford University \\
\texttt{wur@cs.stanford.edu} \\
}

\nipsfinalcopy 

\begin{document}

\maketitle

\begin{abstract}
Our goal is to be able to build a generative model from a deep neural network architecture to try to create music that has both harmony and melody and is passable as music composed by humans. Previous work in music generation has mainly been focused on creating a single melody. More recent work on polyphonic music modeling, centered around time series probability density estimation, has met some partial success. In particular, there has been a lot of work based off of Recurrent Neural Networks combined with Restricted Boltzmann Machines (RNN-RBM) and other similar recurrent energy based models. Our approach, however, is to perform end-to-end learning and generation with deep neural nets alone.
\end{abstract}

\section{Introduction}
\label{intro}
Music is the ultimate language. 
Many amazing composers throughout history have composed pieces that were both creative and deliberate. Composers such as Bach were well known for being very precise in crafting pieces with a great deal of underlying musical structure. Is it possible then for a computer to also learn to create such musical structure?

Inspired by a blog post that was able to create polyphonic music that seemed to have a melody and some harmonization \cite{hexahedria_music}, we decide to tackle the same problem.
We try to answer two main questions

\begin{enumerate}
\item Is there a meaningful way to represent notes in music as a vector? That is, does a method of characterizing meaning of words like word2vec\cite{mikolov} translate to music?
\item Can we build interesting generative neural network architectures that effectively express the notions of harmony and melody? Most pieces have a main melody throughout the piece that it might expand on; can our neural network do the same?
\end{enumerate}

\section{Background and Related Work}
\label{background}
One of the earliest papers on deep learning-generated music, written by Chen et al \cite{chen}, generates one music with only one melody and no harmony. The authors also omitted dotted notes, rests, and all chords. One of the main problems they cited is the lack of global structure in the music.

This suggests that there are two main directions to improve upon
\begin{enumerate}
\item
create music with musical rhythm, more complex structure, and utilizing all types of notes including dotted notes, longer chords, and rests. 
\item
create a model capable of learning long-term structure and possessing the ability to build off a melody and return to it throughout the piece
\end{enumerate}

Liu et al. \cite{liu} tackle the same problem but are unable to overcome either challenge. They state that their music representation does not properly distinguish between the melody and other parts of the piece and in addition do not address the full complexity of most classical pieces. They cite two papers that try to tackle each of the aforementioned problems. 

Eck et al. \cite{eck} use two different LSTM networks -- one to learn chord structure and local note structure and one to learn longer term dependencies in order to try to learn a melody and retain it throughout the piece. This allows the authors to generate music that never diverges far from the original chord progression melody. However, this architecture trains on a set number of chords and is not able to create a more diverse combination of notes. 

On the other hand Boulanger-Lewandowski et al. \cite{boulanger} try to deal with the challenge of learning complex polyphonic structure in music. They use a recurrent temporal restricted Boltzmann machine (RTRBM) in order to model unconstrained polyphonic music. Using the RTRBM architecture allows them to represent a complicated distribution over each time step rather than a single token as in most character language models. This allows them to tackle the problem of polyphony in generated music.

In our project, we will mainly tackle the problem of learning complex structure and rhythms and compare our results to Boulanger-Lewandowski et al.

\section{Data}
\label{data}

One of the primary challenges in training models for music generation is choosing the right data representation. We chose to focus on two primary types: midi files with minimal preprocessing and a "piano-roll" representation of midi files.

\subsection{Midi data}
Midi files are structured as a series of concurrent tracks, each containing a list of meta messages and messages. We extract the messages pertaining to the notes and their duration and encode the entire message as a unique token. 
For example, "note-on-60-0" followed by "note-off-60-480" would translate into two separate messages or tokens. Together, these two messages would instruct a midi player to play "middle-C" for 480 ticks, which translates to a quarter note for most midi time scales. We flatten the tracks so that the tokens of the separate tracks of a piece would be concatenated end-to-end.

We started by downloading the entire Bach corpus from MuseData \footnotemark because Bach was comparatively the most prolific composer on that website. In total, there were 417 pieces for 1,663,576 encoded tokens in our Bach corpus. We also made sure to normalize the ticks per beat for each piece. We did not, however, transpose every piece into the same key, which has been shown to improve performance \cite{boulanger}.

\footnotetext{The data for the Bach corpus was pulled directly from the MuseData website at \url{http://musedata.org/}.}

\begin{figure}[ht]
\begin{center}
\includegraphics[width=0.4\linewidth]{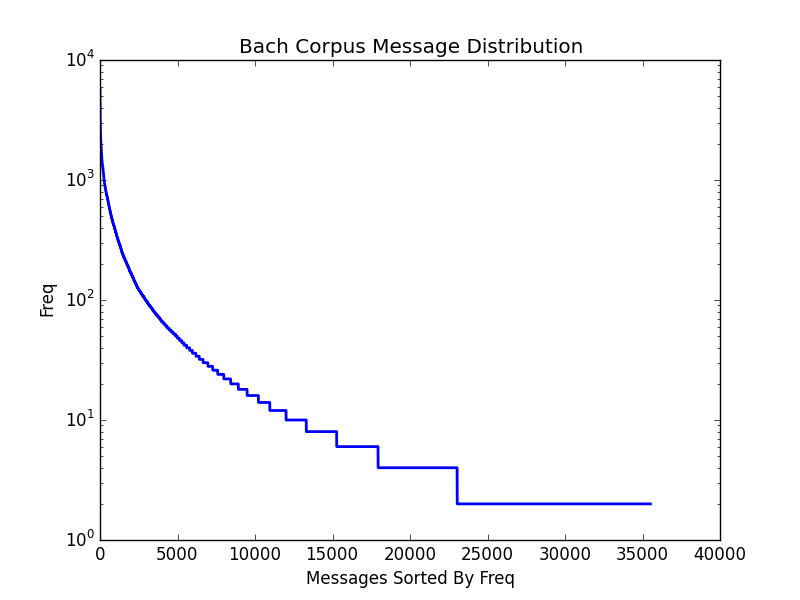}
\includegraphics[width=0.4\linewidth]{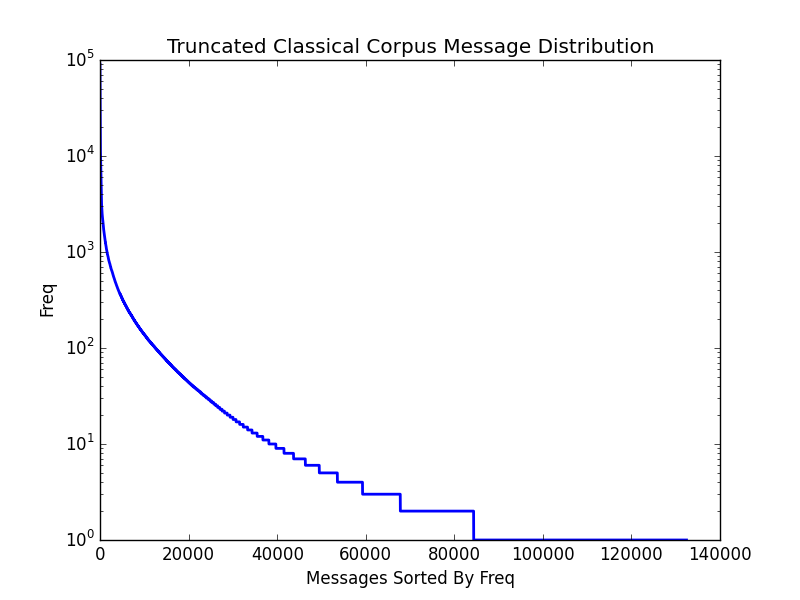}
\end{center}
\caption{Message or token distribution for both the Bach only corpus (left) and for the truncated version of the classical music corpus (right).\label{fig:tokens}}
\end{figure}

Furthermore, we scraped additional midi files from other online repositories\footnotemark that had a mix of different classical composers. This expanded our corpus from around 1 million tokens to around 25 million tokens. Due to memory constraints on our model, we primarily operated on a truncated version of this dataset that contained 2000 pieces.

\footnotetext{The sites we scraped additional midi files from include \url{http://classicalmidiresource.com/}, \url{http://midiworld.com/}, and \url{http://piano-midi.de/}}

\begin{center}
\begin{tabular} {c c c c}
Corpus & Words & Unique Tokens \\ 
\hline
Bach Only & 1,663,576 & 35,509 \\ 
Full Classical & 24,654,390 & 175,467 \\ 
Truncated Classical & 11,413,884 & 132,437 \\  
\end{tabular}
\end{center}

We also compared the token distribution for both the Bach only midi corpus and the entire classical midi corpus as seen in figure \ref{fig:tokens}. We see that there are many messages with very low frequency in both datasets. Indeed, for both datasets more than two-thirds of the unique tokens occurred less than 10 times. 

More importantly however, the drawback of encoding midi messages directly is that it does not effectively preserve the notion of multiple notes being played at once through the use of multiple tracks. Since we concatenate tracks end-to-end, we posit that it will be difficult for our model to learn that multiple notes in the same position across different tracks can really be played at the same time.

\subsection{Piano roll data}
In order to address the drawbacks outlined above, we turn to a different data representation. Instead of having tokens split by track, we represent each midi file as a series of time steps where each time step is a list of note ids that are playing.

\begin{figure}[ht]
\begin{center}
\includegraphics[width=0.4\linewidth]{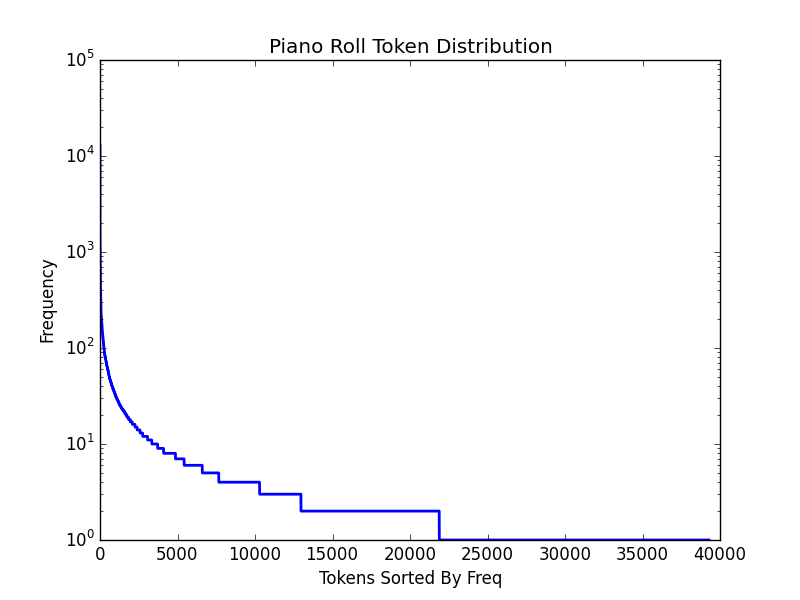}
\end{center}
\caption{Frequency distribution of all the tokens in the 'Muse-All' piano roll dataset.}
\end{figure}

We retrieved the piano roll representation of all the pieces on MuseData from \footnotemark. The dataset was created by sampling each midi file at eighth note intervals; the pieces were also transposed to C-Major/C-minor. The training set provided had 524 pieces for a total of 245,202 time steps. We encode each time step by concatenating the note ids together to form a token (e.g. a C-Major chord would be represented as "60-64-67"). Furthermore, as we were concerned about the number of unique tokens, we randomly chose 3 notes if the polyphony exceeded 4 at any particular time step.

\footnotetext{The MuseData piano roll dataset is available on Boulanger-Lewandowski's website at \url{http://www-etud.iro.umontreal.ca/~boulanni/icml2012}.}

\begin{center}
\begin{tabular} {c c}
Dataset & Unique Tokens \\
\hline
Muse-All & 39,289 \\
Muse-Truncated & 21,510 \\
\end{tabular}
\end{center}

\section{Approach}
\label{approach}
We use a 2-layered Long Short Term Memory (LSTM) recurrent neural network (RNN) architecture to produce a character level model to predict the next note in a sequence.

In our midi data experiments, we treat a midi message as a single token, whereas in our piano roll experiment, we treat each unique combination of notes across all time steps as a separate token.

We create an embedding matrix which maps each token into a learned vector representation. A sequence of tokens in a piece is then concatenated into a list of embedding vectors that forms the time sequence input that is fed into the LSTM.

The output of the LSTM is fed into softmax layer over all the tokens. The loss corresponds to the cross entropy error of our predictions at each time step compared to the actual note played at each time step.

Our architecture allows the user to set various hyperparameters such as number of layers, hidden unit size, sequence length, batch size, and learning rate. We clip our gradients to prevent our gradients from exploding. We also anneal our learning rate when we see that the rate that our training error is decreasing is slowing.

We generate music by feeding a short seed sequence into our trained model. We generate new tokens from the output distribution from our softmax and feed the new tokens back into our model. We used a combination of two different sampling schemes: one which chooses the token with maximum predicted probability and one which chooses a token from the entire softmax distribution.
We ran our experiments on AWS g2.2xlarge instances. Our deep learning implementation was done in TensorFlow.

\section{Experiments}
\label{experiments}
\subsection{Baseline}

\begin{figure}[ht]
\begin{center}
\includegraphics[width=0.8\linewidth]{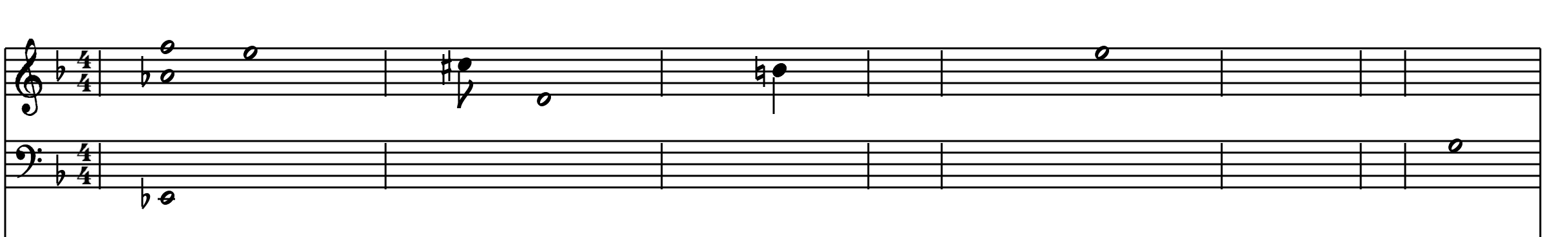}
\includegraphics[width=0.8\linewidth]{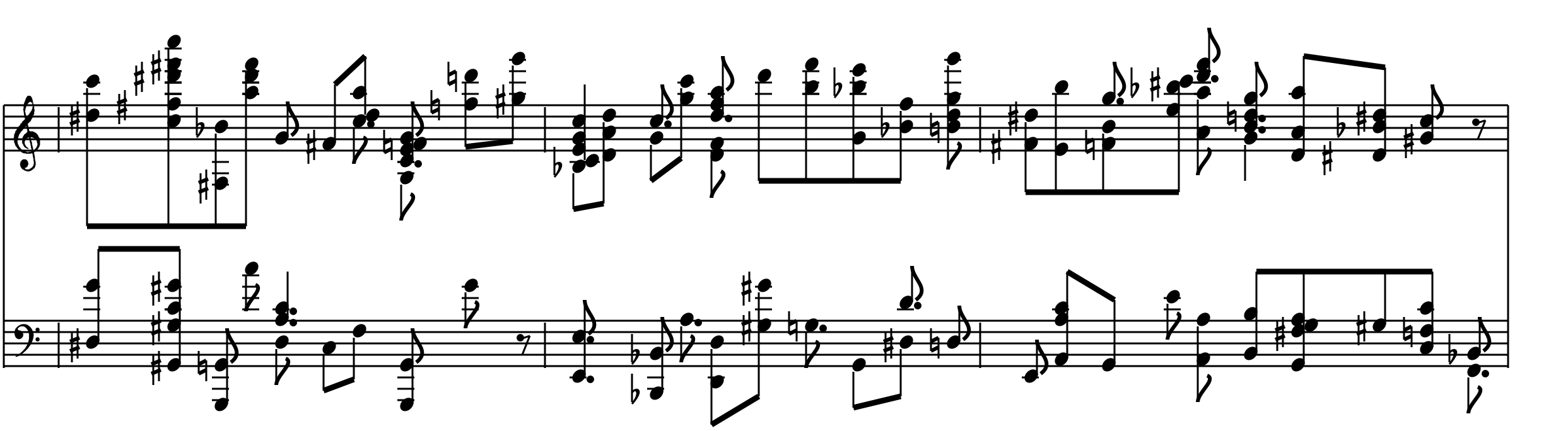}
\end{center}
\caption{(Top): Generated baseline midi files from an untrained model. (Bottom): Weighted sample of tokens from the piano roll representation. \label{baselines}}
\end{figure}

For our midi baseline, we had our untrained model generate sequences. As you can see in figure \ref{baselines}, our model was not able to learn the "on-off" structure of the midi messages, which results in many rests. For our piano roll baseline, we sample random chords from our piano roll representation weighted by how frequent they occur in our corpus. 

We see that for the piano roll the music is very dissonant, and while each chord may sound reasonable, there is no local structure from chord to chord.

\subsection{Bach midi experiment}
We first train our model on the "Bach Only" midi dataset. We trained for around 50 epochs, which took about 4 hours to train on a GPU.

\begin{center}
\begin{tabular} {c l}
Hidden State & 128 \\
Token Embedding Size & 128 \\
Batch Size & 50 \\
Sequence Length & 50 \\
\end{tabular}
\end{center}

\begin{figure}[ht]
\begin{center}
\includegraphics[width=0.8\linewidth]{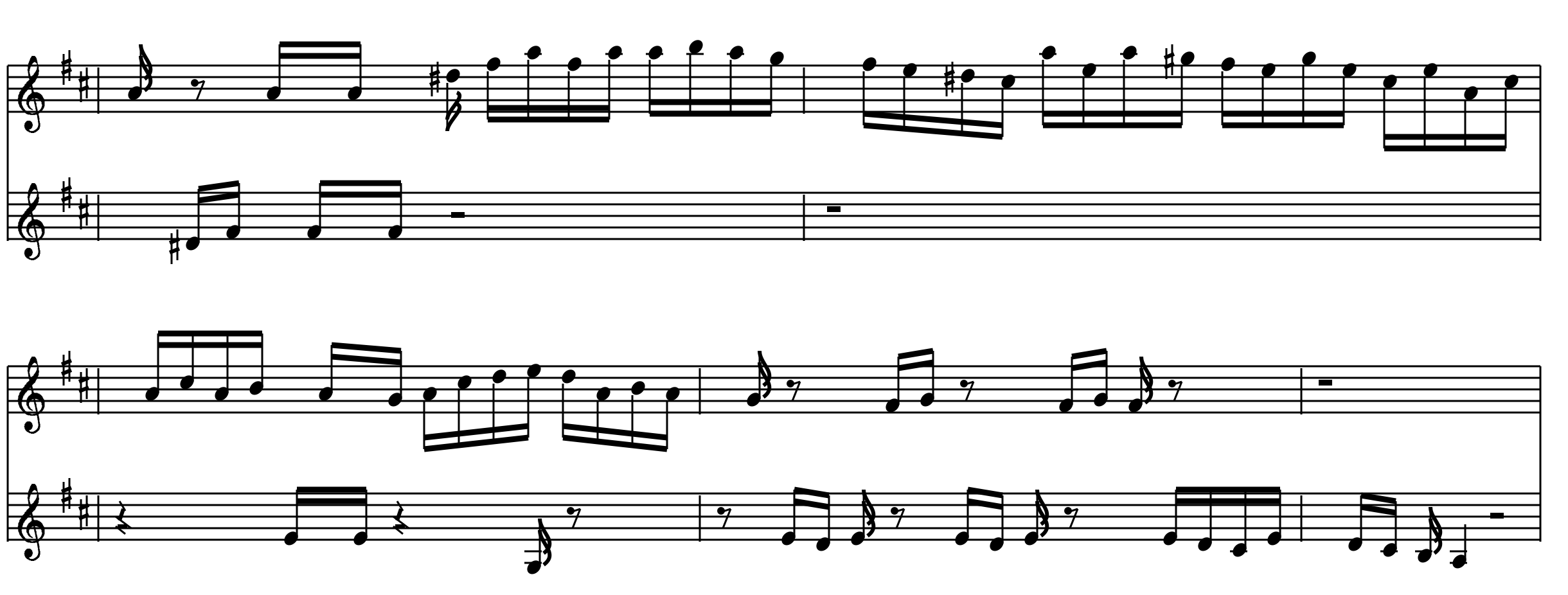}
\end{center}
\caption{Music generated from 'Bach Only' dataset.}
\end{figure}

\subsection{Classical midi experiment}
We use the same architecture as in the Bach-Midi experiment on the "Truncated Classical" dataset due to time constraints. 15 epochs took 22 hours on a GPU. Furthermore, due to limitations on device memory on AWS's g2.2xlarge, we were forced to reduce the batch size and the sequence length.

\begin{center}
\begin{tabular} {c l}
Hidden State & 128 \\
Token Embedding Size & 128 \\
Batch Size & 25 \\
Sequence Length & 25 \\
\end{tabular}
\end{center}

\begin{figure}[ht]
\begin{center}
\includegraphics[width=0.8\linewidth]{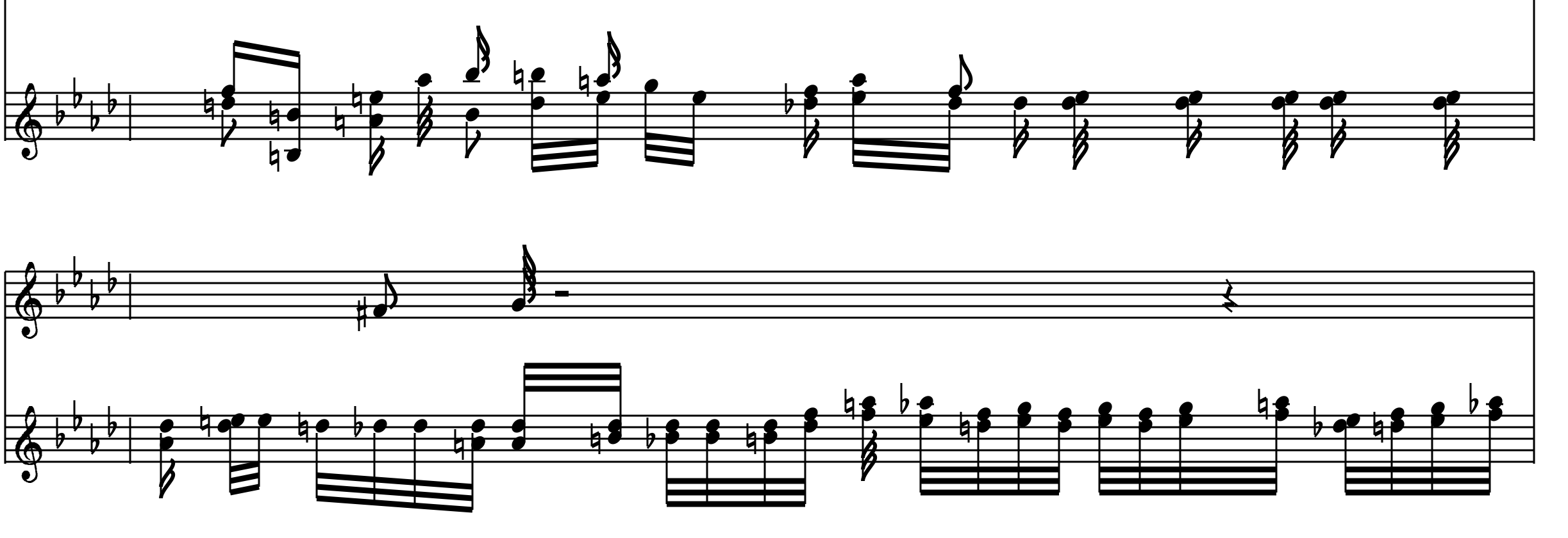}
\end{center}
\caption{Music generated from the 'Truncated Classical' dataset.}
\end{figure}

\subsection{Discussion}
Interestingly, we found that the sequences produced by the model trained on the "Bach Only" data were more aesthetically pleasing than the one trained on a grab bag of different classical pieces. We believe that relative size of the character set of the classical midi model relative to the Bach model severely hindered its ability to learn effectively.

\begin{figure}[h]
\begin{center}
\includegraphics[width=\linewidth]{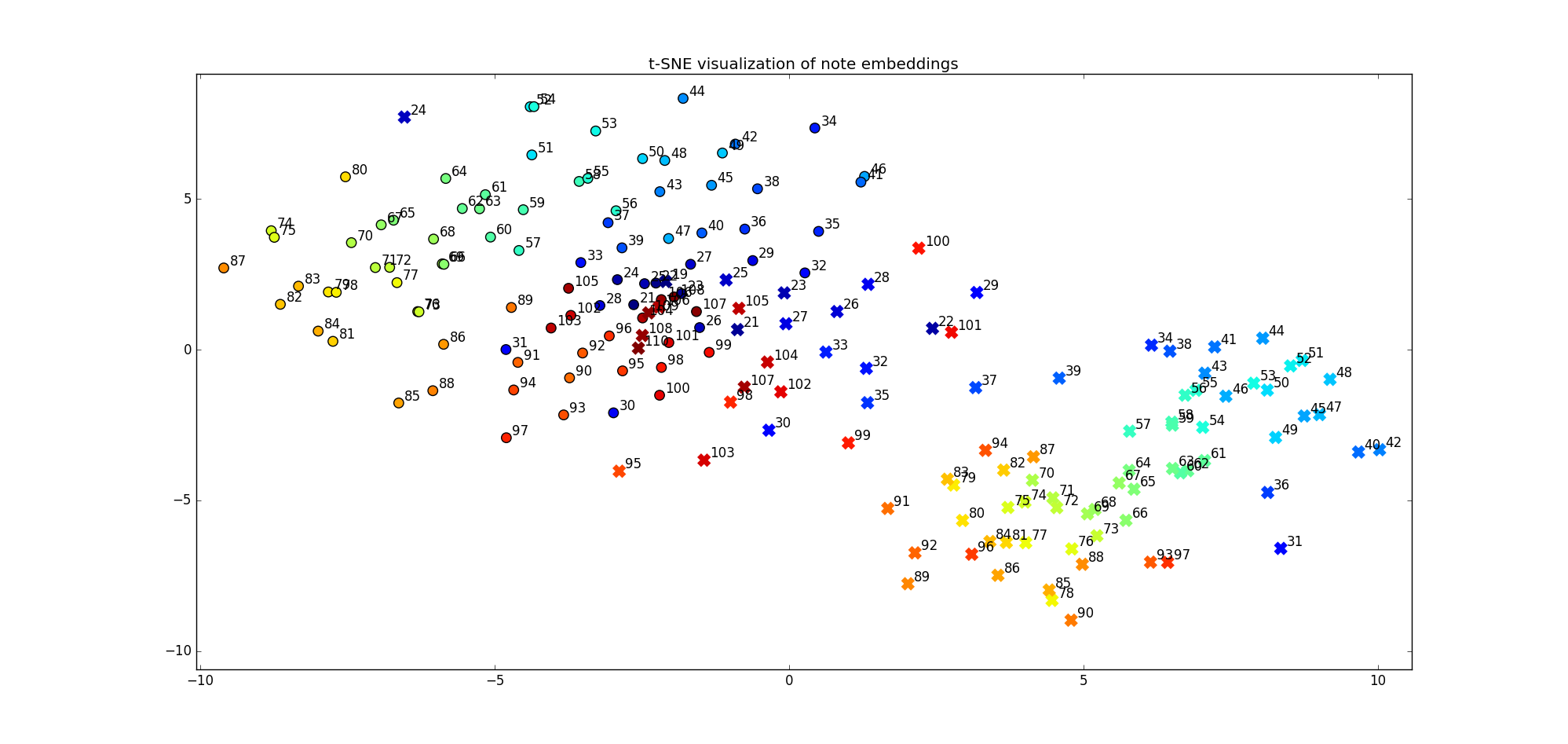}
\end{center}
\caption{t-SNE visualization of embedding vectors from the classical midi experiment.\label{fig:tsne_midi}}
\end{figure}

We also use t-SNE to visualize our embedding vectors for our character model as a measure of success. We can see the results in figure \ref{fig:tsne_midi}.
The circles denote midi ON messages while the x's represent midi OFF messages. The numbers represent midi note ids (lower numbers represent lower frequency), which are also color-coded from blue to red (low to high respectively).

Since we have so many tokens in our model, we filter our visualization to only show notes that were played for 60 ticks.

Note that there are clear clusters between the on and off messages for the medium frequency notes (the notes that are played most often), while the  the rare low and high notes are clumped together in an indistinct cloud in the center.

In addition, the model seems to learn to group similar pitches close together and to have some sort of linear progression from low pitches to high pitches. The on notes and off notes both have a general pattern of lower pitches in the top right to higher pitches in the bottom left. 

\subsection{Piano roll experiment}
We ran this experiment with the same parameters as the "Bach-Midi Experiment." We ran it with for 800 epochs, which took 7 hours on a AWS g2.2xlarge instance. We also ran the same configuration on the truncated dataset for 100 epochs, which took 7 hours on a CPU. 

\begin{figure}[ht]
\begin{center}
\includegraphics[width=0.8\linewidth]{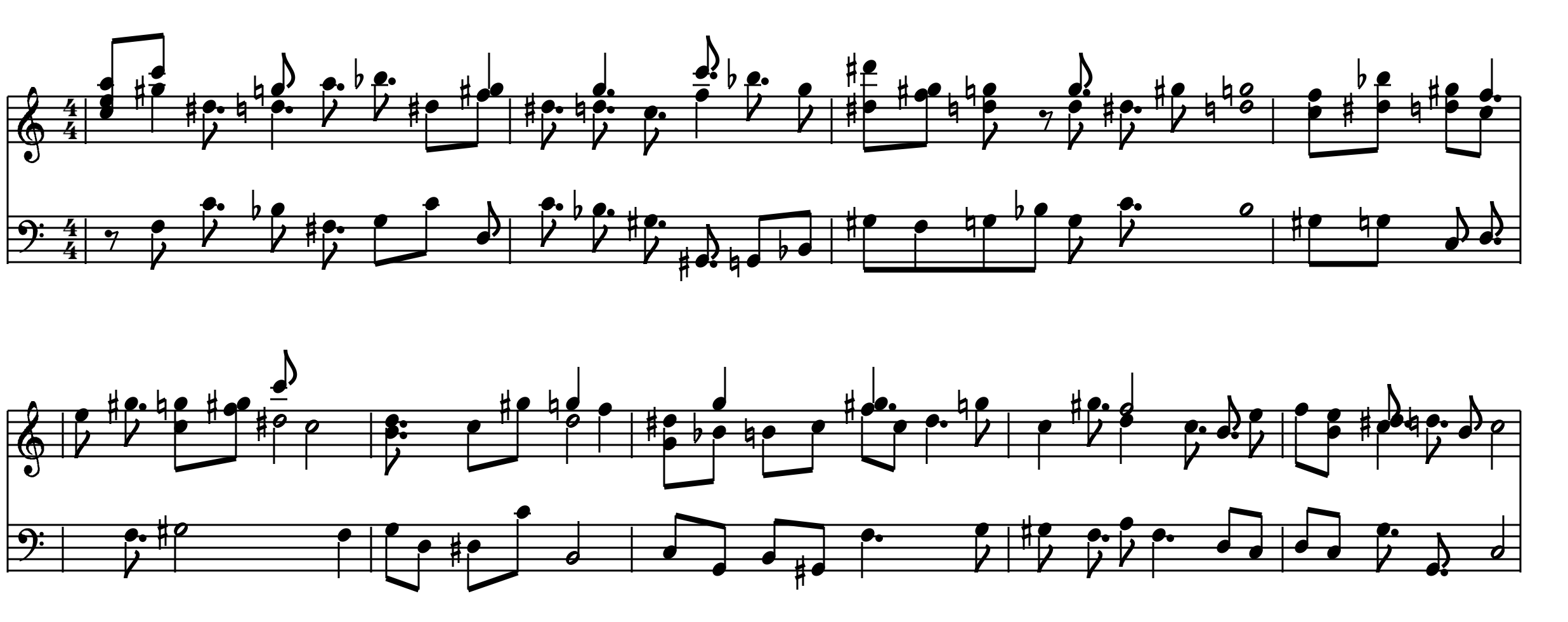}
\end{center}
\begin{center}
\includegraphics[width=0.8\linewidth]{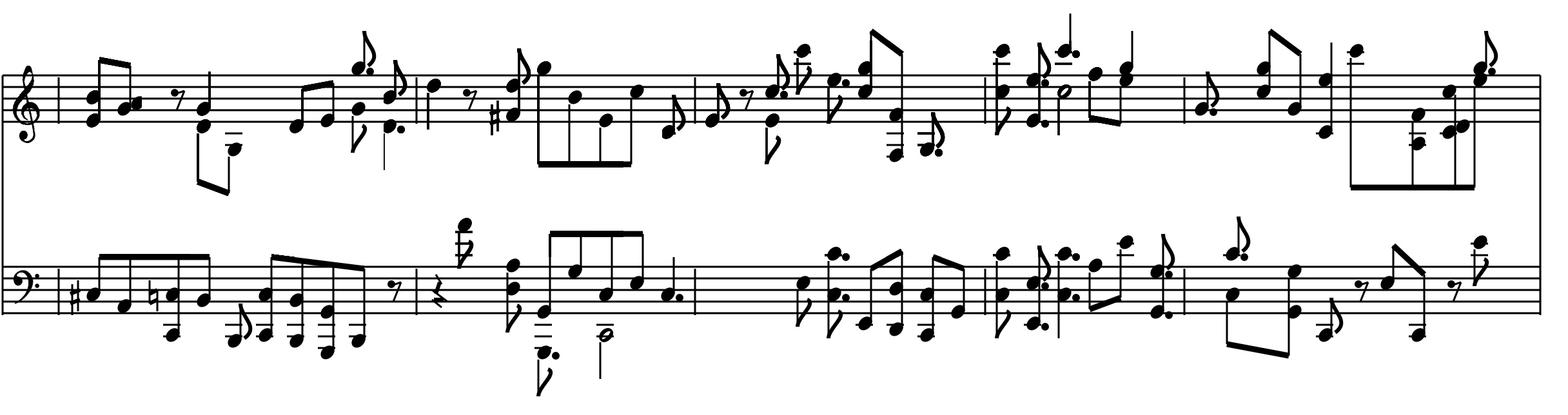}
\end{center}
\caption{Music generated from the Muse piano roll data. The top 4 lines are from the 'Muse-All' dataset and the last two lines are from the 'Muse-Truncated' dataset.}
\end{figure}

\begin{figure}[h]
\begin{center}
\includegraphics[width=0.6\linewidth]{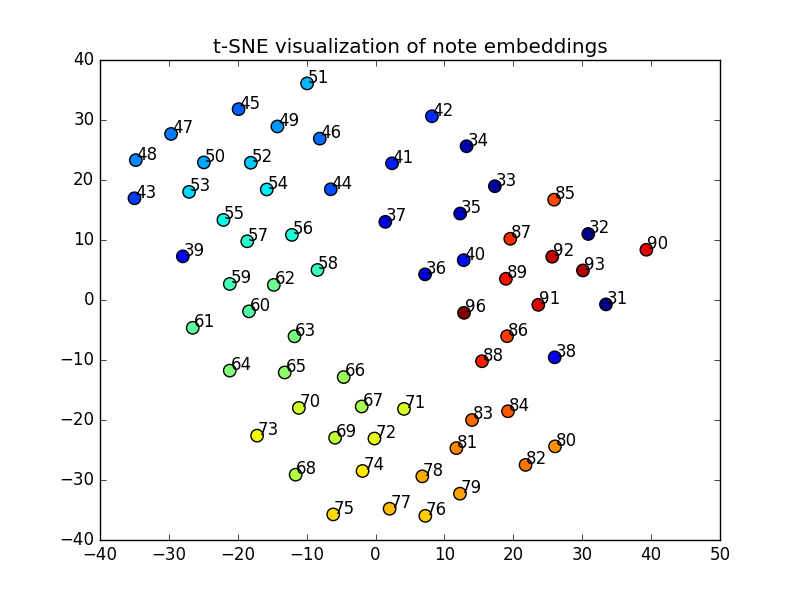}
\end{center}
\caption{t-SNE visualization of single note embedding vectors from the piano roll experiment. \label{fig:tsne_pianoroll}}
\end{figure}

We again use t-SNE to visualize our embedding vectors for our new model as seen in figure \ref{fig:tsne_pianoroll}.
In this model, since we have tens of thousands of different combinations of notes, we first look at the tokens that encode a single note being played. 

Here we see a similar result where the model is able to disambiguate among the lower and higher pitches cleanly. We again see that the lowest and highest pitches however are also grouped together and are slightly separated from the other note embeddings.

\section{Evaluation}

One of the major challenges of evaluating the quality of our model was incorporating the notion of musical aesthetic. That is, how "good" is the music that our model ultimately generates? As such, we devised a blind experiment where we asked 26 volunteers to offer their opinion on 3 samples of generated music.
\begin{itemize}
\item We asked them to hear the 3 samples back-to-back.
\item We asked them to rate on a scale from 1 to 10.
\begin{itemize}
	\item A 1 rating would be "completely random noise"
    \item A 5 rating would be "musically plausible"
    \item A 10 rating would be "composed by a novice composer"
\end{itemize}
\end{itemize}

The identify of the samples was as follows:

\begin{tabular} {c | l}
Sample 1 & 10 second clip of the "Bach Midi" model \\
Sample 2 & 16 second clip of the "7\_RNN-NADE sequence" from \cite{boulanger}.\footnotemark\\
Sample 3 & 11 second clip of the "Piano roll" model trained on the "Muse-All" dataset
\end{tabular}

\footnotetext{This sequence can be downloaded from \url{http://www-etud.iro.umontreal.ca/~boulanni/icml2012}. Click the "MP3 samples" link.}

We chose to compare our sequences with a RNN-Neural Autoregressive Distribution Estimator (RNN-NADE) sequence from \cite{boulanger} because it achieved similar results as other commonly used techniques such as RNN-RBM and RTRBM and is robust as a distribution estimator\cite{boulanger}.

\begin{figure}[ht]
\begin{center}
\includegraphics[width=0.49\linewidth]{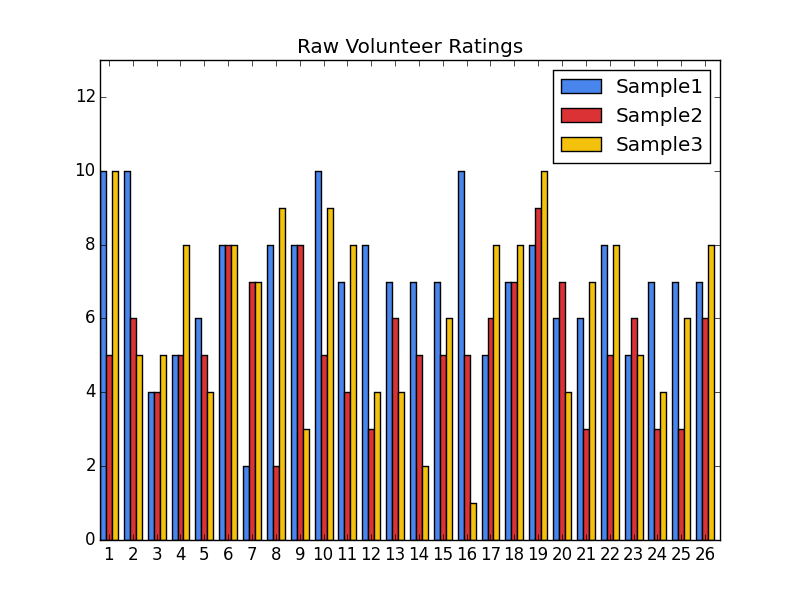}
\includegraphics[width=0.49\linewidth]{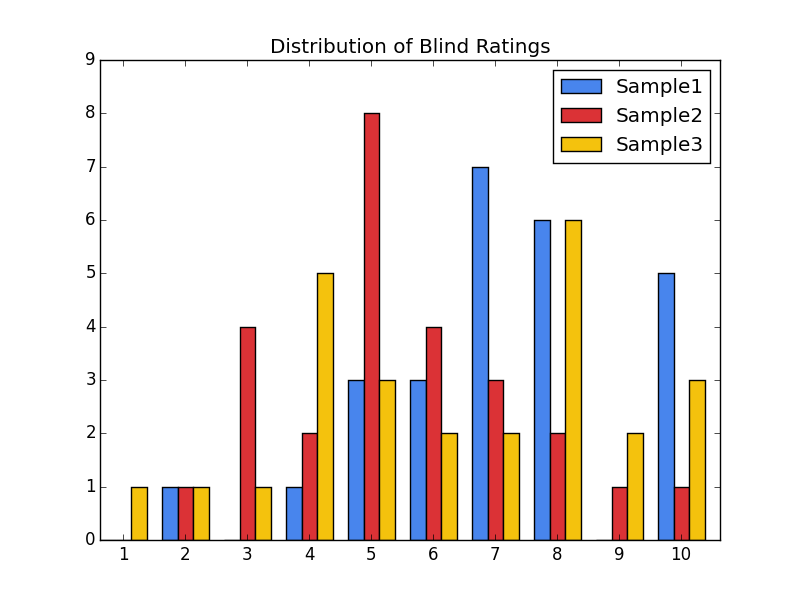}
\end{center}
\caption{(Left): Raw voting values for each sequence for 26 volunteers. (Right): Histogram of ratings.\label{fig:raw-volunteer-ratings}}
\end{figure}

Our results indicate that our models did in fact produce music that is at least comparable in aesthetic quality to the RNN-NADE sequence. Indeed, in figure \ref{fig:raw-volunteer-ratings}, we see that only 3 out of the 26 volunteers said that they liked the sequence from the RNN-NADE better. (An additional 3 said that they liked it just as much as one of our sequences.) That being said, in the histogram in figure \ref{fig:raw-volunteer-ratings}, we see that the samples had an average rating of 7.0$\pm$1.87, 5.3$\pm$1.7, and 6.2$\pm$2.4 respectively, which suggests that our sample size was too small to distinguish the different samples statistically. 

\section{Conclusion and Future Work}
We were able show that a multi-layer LSTM, character-level language model applied to two separate data representations is capable of generating music that is at least comparable to sophisticated time series probability density techniques prevalent in the literature. We showed that our models were able to learn meaningful musical structure.

This paper's writing comes at an interesting time in the space of deep learning generated art. In the last week, Google has announced its new Magenta program\footnotemark, a TensorFlow-backed machine learning platform for generating art. Google also released a 90-second clip computer-generated melody with an accompanying drum line. 
\footnotetext{Blog post introducing Magenta can be found here: \url{http://magenta.tensorflow.org/welcome-to-magenta}}

Given the recent enthusiasm in machine learning inspired art, we hope to continue our work by introducing more complex models and data representations that effectively capture the underlying melodic structure. Furthermore, we feel that more work could be done in developing a better evaluation metric of the quality of a piece -- only then will we be able to train models that are truly able to compose original music!

\label{conclusion}

\bibliography{music}{}
\bibliographystyle{plain}

\end{document}